\pdfoutput=1
\documentclass[10pt, a4paper]{article}
\usepackage{lrec-coling2024} % this is the new style
\usepackage{times}
\usepackage{latexsym}
\usepackage[T1]{fontenc}
\usepackage[utf8]{inputenc}
\usepackage{microtype}
\usepackage{inconsolata}
\usepackage{url}
\usepackage{amsmath,amssymb}

\usepackage{amsfonts}
\usepackage{helvet}
\usepackage{courier}
\usepackage{url}
\usepackage{float,color}
\usepackage{times}
\usepackage{algorithm}
\usepackage{algorithmic}
\usepackage{float}
\usepackage{pdfpages}
\usepackage{changes}
\usepackage{epsfig}
\usepackage{stfloats}
\usepackage{url}
\usepackage{diagbox}
\usepackage{subfigure}
\usepackage{epstopdf}
\usepackage{multicol}
\usepackage{makecell}
\usepackage{longtable}
\usepackage{dsfont,amsfonts,color}
\usepackage{multirow}
\usepackage{booktabs}
\usepackage{xr}
\usepackage{tablefootnote}
\def\p{{\bf p}}

\def\y{{\bf y}}

\def\0{{\bf 0}}
\def\1{{\bf 1}}
\def\AM{{\mathcal A}}

\def\CM{{\mathcal C}}

\def\MM{{\mathcal M}}

\def\SM{{\mathcal S}}

\def\YM{{\mathcal Y}}

\graphicspath{{img/}}

\title{On the use of Silver Standard Data for Zero-shot Classification Tasks in Information Extraction}

\name{Jianwei Wang$^{1}$, Tianyin Wang$^{2}$, Ziqian Zeng$^{1}$} 

\address{$^{1}$Shien-Ming Wu School of Intelligent Engineering, South China University of Technology, China, \\ $^{2}$School of Computer Science \& Engineering, South China University of Technology, China \\
         wiwjwilliam@mail.scut.edu.cn, cstianyinwang@mail.scut.edu.cn, zqzeng@scut.edu.cn\\ \\}

\abstract{
The superior performance of supervised classification methods in the information extraction (IE) area heavily relies on a large amount of gold standard data. 
Recent zero-shot classification methods converted the task to other NLP tasks (e.g., textual entailment) and used off-the-shelf models of these NLP tasks to directly perform inference on the test data without using a large amount of IE annotation data. 
A potentially valuable by-product of these methods is the large-scale silver standard data, i.e., pseudo-labeled data by the off-the-shelf models of other NLP tasks. 
However, there is no further investigation into the use of these data. 
In this paper, we propose a new framework, Clean-LaVe, which aims to utilize silver standard data to enhance the zero-shot performance.
Clean-LaVe includes four phases: (1) Obtaining silver data; (2) Identifying relatively clean data from silver data; (3) Finetuning the off-the-shelf model using clean data; (4) Inference on the test data.
The experimental results show that Clean-LaVe can outperform the baseline by 5\% and 6\% on TACRED and Wiki80 dataset in the zero-shot relation classification task, and by 3\% \textasciitilde 7 \% on Smile (Korean and Polish) in the zero-shot cross-lingual relation classification task, and by 8\% on ACE05-E+ in the zero-shot event argument classification task. The code is share in \href{https://github.com/wjw136/Clean_LaVe.git}{https://github.com/wjw136/Clean\_LaVe.git}. 
 \\ \newline  }

\begin{document}

\maketitleabstract

\section{Introduction}
Information Extraction (IE) is a fundamental problem in natural language processing. 
The predominant approaches to solve IE tasks are supervised methods \cite{shen2022parallel, zhu2022boundary, zhong2021frustratingly, lyu2021relation, yang2019exploring, lu2023event}. 
Supervised methods require a large amount of gold standard data, which restricts their applications to real-world scenarios where large-scale annotated data are not available. 
Zero-shot methods \cite{lyu2021zero,sainz2021label} have been proposed to alleviate this issue. 
We focus on zero-shot classification tasks in IE such as relation extraction (RE), cross-lingual relation extraction, and event argument classification (EAC). 
Zero-shot (cross-lingual) RE aims to identify the semantic relation between two entities in unstructured texts without using any annotated RE data (in the target language).
Zero-shot EAC aims to assign roles to argument spans using any annotated EAC data.

Recent works \cite{sainz2021label,saintz2022textual, sainz2022zs4ie,lu2022summarization} attempt to convert the zero-shot RE task and EAC task to other NLP tasks and used off-the-shelf models of these tasks to infer the relation types without using a large amount of RE or EAC annotated data. 
\citet{sainz2021label} used a well-trained textual entailment (TE) model to directly infer relation types on the RE test data by converting a RE task to a TE task.
Their subsequent work \cite{saintz2022textual} also used a TE model to infer argument roles on the test data by converting an EAC task to a TE task. 
This series of work is named LaVeEntail.
SURE \cite{lu2022summarization} formulated a RE task to a summarization task, and used a small amount of RE annotated data to finetune a well-trained summarization model thus it can perform inference on the RE test data. 
We term the TE model and summarization model in the above methods as pre-trained models. 
The concept of pretraining derives from transfer learning \cite{pan2009survey}. 
A model is first pre-trained on the source task, i.e., textual entailment or summarization, and then finetuned on the target task, i.e., relation extraction and event argument classification.

Since pre-trained models can directly infer the categories of unlabeled data, they can serve as low-cost annotators, producing large-scale silver standard data.
However, in the above works, silver standard data are not well-exploited.
The straightforward way to utilize them is to directly train a supervised classifier on silver standard data. However, the performance is usually unsatisfactory due to the noisy nature of silver standard data. 
Learning with noisy labels has been well studied in the literature \cite{frenay2013classification,algan2021image,bo2020survey}. 
One direction is to develop noise-robust losses that can mitigate the effect of noisy labels \cite{ghosh2017robust,zhang2018generalized,charoenphakdee2019symmetric,kim2019nlnl,lyu2019curriculum,menon2020can,thulasidasan2019combating}.
Another direction is to identify noisy data or clean data and deal with them separately either by re-weighting or converting to a semi-supervised learning task. \cite{han2018co,jiang2018mentornet,arazo2019unsupervised,kim2019nlnl,shu2019meta,yao2019searching,li2020dividemix}. 
The setting of traditional noisy labels learning does not consider the existence of a pre-trained model. 
Is there a better way to utilize silver standard data when a pre-trained model is available?
According to our best knowledge, there is no further investigation on the use of potentially valuable silver standard data when there exists a pre-trained model. 

\begin{figure*}[htbp]

\centering

\includegraphics[width=0.9\textwidth]{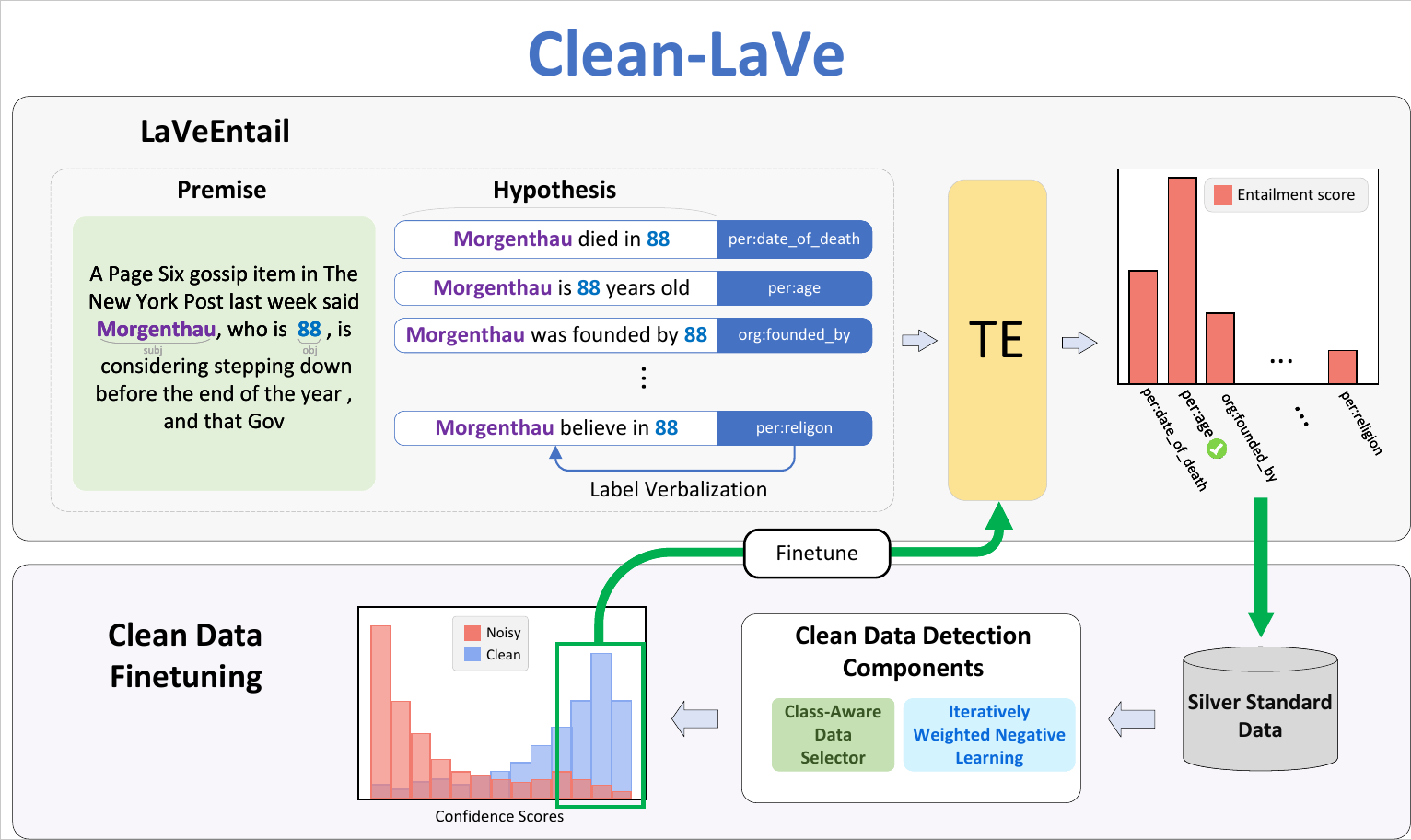}
\caption{
The diagram shows the procedure of Clean-LaVe in the zero-shot relation extraction task.
First, we apply LaVeEntail on an unlabeled dataset, obtaining standard silver data. 
The clean data detection module uses confidence scores to distinguish the clean and noisy samples. 
The selected clean data are used to finetune the TE model.
Finally, we use this TE model to infer relation types on the test set. 
% \revisezq{check it . it is old. no mention weighted negative learning}
}
% \revisejw{
% Clean data detection module just contains both weighted NL and class-aware selector and maybe i can just describe both of them as a whole named clean data detection component?
% }
\label{fig:procedure}
\end{figure*}

In this paper, we propose a novel framework called \textbf{Clean-LaVe}. The framework involves two main steps: firstly, detecting a small subset of clean data from the silver standard data using the clean data detection module, and secondly, utilizing the selected clean data to finetune the pre-trained model. 
The overall procedure is illustrated in Figure \ref{fig:procedure}.

Within the clean data detection module, we introduce a \textbf{iteratively weighted negative learning} algorithm to obtain confidence scores that allow us to distinguish clean data from noisy data. 
The original negative learning algorithm \cite{kim2019nlnl} only performs well when the dataset is balanced \cite{huang2022uncertainty, lu2023label}. 
However, in real-world scenarios, this assumption may not hold. 
To address this issue, we introduce an iterative weighting strategy to allow the algorithm to handle an imbalance dataset.

To select clean data, confidence scores serve as a straightforward metric \cite{kim2019nlnl}. 
However, data from certain classes possibly have high confidence scores while others yield low scores. 
In such cases, selecting data solely based on confidence scores may lead to a narrow range of classes being selected, potentially harming overall performance. 
To mitigate this issue, we develop a \textbf{class-aware data selector} that enables the selection of data from a broader range of classes.

%To obtain silver standard data, we use the pre-trained model to annotate the unlabeled RE training data.
%The distribution of silver standard data is the same as the RE test data. 
%However, unlabeled data of the same distribution with test data are still scarce while unlabeled data of different distribution are common.
%Hence, we use the finetuned model to annotate some unlabeled data of different distributions, i.e., only partial classes are overlapped with the test data. 
%The experimental results show that our method can utilize the silver standard data of different distribution, here called \textbf{Extra Silver Data}, to further improve the performance. 

Clean-LaVe is a general framework that can be used in scenarios where a pre-trained model serves as an annotator.  
In our experiments, we demonstrate the usability of Clean-LaVe in various zero-shot classification tasks, such as zero-shot RE, zero-shot cross-lingual RE, and zero-shot EAC. 
For these aforementioned zero-shot tasks, we utilize a TE model as the pre-trained model to acquire silver standard data.

Our contributions are summarized as follows,

$\bullet$ We propose Clean-LaVe to first detect a small amount of clean data which are later used to finetune the pre-trained model. We then use the finetuned model to infer the categories on the test data.

$\bullet$ We propose a clean data detection module that enhances the selection process through Iteratively Weighted Negative Learning and Class-Aware Data Selector.

$\bullet$ The experimental results demonstrate that our method can outperform the baseline by a large margin on various zero-shot classification tasks.

% \href{https://anonymous.4open.science/r/Clean\_LaVe-3DE3}{https://anonymous.4open.science/r/Clean\_LaVe-3DE3}.

% By using additional silver standard data of different distributions, the performance can be further improved. 
% Experimental results show that Clean-LaVe can be generalized other zero-shot tasks such as zero-shot cross-lingual RE, zero-shot event detection, and zero-shot event argument classification tasks. 
%The experiments also reveal that Clean-LaVe can be generalized to other zero-shot tasks, including zero-shot cross-lingual RE, zero-shot event detection, and zero-shot event argument classification tasks.

\section{Related Work}

\paragraph{\textbf{Zero-shot Relation Extraction.}}
In classical zero-shot learning settings, the classes in the training and test phases are disjoint.   
In the training phase, it requires a large amount of annotated RE data from seen classes.  
In the test phase, zero example for each unseen relation type during the test phase is needed. 
Recent works \cite{levy2017zero, obamuyide2018zero,zhao2023re_matching} formulated the zero-shot RE to other NLP tasks such as reading comprehension \cite{levy2017zero,zhao2023re_matching} and textual entailment \cite{obamuyide2018zero}.
%Recent works \cite{han2018fewrel,soares2019matching,chen2021zs}.
%\citet{levy2017zero} reformulated the zero-shot slot filling task to a reading comprehension task. 
%The slot filling task aims to predict the entity given a relation type and another entity. 
%\citet{obamuyide2018zero} reduced the zero-shot and few-shot relation extraction task to a textual entailment problem. 
%The premise is the sentence containing two entities. 
%The hypothesis is a relation description template instantiated by two entities. 

However, the classical zero-shot setting still requires a large amount of annotated data in the training phase. 
Recent works \cite{goswami2020unsupervised,sainz2021label,tran2021one,lu2022summarization,rahimi2023improving} push the zero-shot setting to an extreme case where annotated data is not available in the training phase. 
They obtained supervision from other available resources such as language models \cite{tran2021one, zhang2023QA4RE}, relation descriptions, and off-the-shelf models from other NLP tasks. 
%\cite{goswami2020unsupervised} reformulated the zero-shot slot filling task to a cloze question answering problem.
%LaVeEntail \cite{sainz2021label} utilized an off-the-shelf textual entailment model to directly infer RE data on the test set. 
%\cite{tran2021one} compared the similarity of the embeddings of manually created relation exemplars and the input sentence to infer the relation types.
%SURE \cite{lu2022summarization} converted the few-shot RE task to a summarization task. 
QA4RE \cite{zhang2023QA4RE} aligns RE with question answering.
QA4IE \cite{zhang2023QA4RE} is the state-of-the-art method in the zero-shot RE task, primarily owing to the powerful capacity of Large Language Models. 
In this paper, we focus on this extreme zero-shot setting.

%\paragraph{\textbf{Textual Entailment}}
%Recent work has demonstrate textual entailment to be a viable proxy for text classification task, such as topic and emotion analysis \cite{yin2019benchmarking,sainz2021ask2transformers,zhong2021adapting}. 
%The process of reframing a text classification problem as an entailment task begins with the definition of templates to describe each class label, referred to as the verbalization of a hypothesis. 
% \revisezq{Pre-trained model finetuned on TE dataset, such as MNLI\cite{williams2017broad}, SNLI\cite{bowman2015large}, ANLI\cite{nie2019adversarial}, XNLI\cite{conneau2018xnli}, is employed to determine entailment of these hypothesises. 
% Additionally, there is a pre-trained model\cite{he2021debertav3} finetune on a multilingual NLI dataset\cite{laurer_less_2022} emcompassing 26 languages, making it adaptable for multilingual textual entailment tasks.} 
%\cite{wang2021entailment} demonstrate that textual entailment serves as a robust language understanding task, enabling the transfer of knowledge to other similarly-framed tasks.

\paragraph{\textbf{Zero-shot Cross-lingual Information Extraction.}}
Existing approaches to zero-shot cross-lingual Information Extraction (IE) can be categorized into three main types: translation-based \cite{lou2022translation}, feature-based \cite{huang2022multilingual, ma2023shine}, and distillation-based methods \cite{wu2020single, ma2022wider}.
%\cite{lou2022translation} employed translated dataset to improve training in target languages, but this approach may introduce noise during the translation process.  
%\cite{huang2022multilingual, ma2023shine} devised language-agnostic features to directly train language models.
%\cite{ma2022wider, wu2020single} have leveraged distillation techniques to transfer knowledge from source languages to target languages. 
However, all of these methods require significant manual effort to obtain labeled data for the source languages, which could potentially be replaced by readily available resources in the target language domain, such as off-the-shelf TE models.

\paragraph{\textbf{Zero-shot Event Argument Classification.}} Exiting zero-shot event argument classification tasks are based on label representations \cite{huang2017zero,zhang2021labelaware}, reading comprehension \cite{liu2020event, lyu2021zero, mehta2022improving}, and pre-trained language models \cite{huang2022multilingual, lin2023constraintsEAC}. 
\citet{lin2023constraintsEAC} is the state-of-the-art zero-shot EAC method, which prompts the pre-trained language models and regularizes the prediction by global constraints.

\paragraph{\textbf{Learning with Noisy Labels.}}
One direction of learning with noisy labels is to develop noise-robust loss. 
The widely-used cross entropy (CE) loss in classification tasks has been shown to be not robust against label noise \cite{ghosh2017robust}. 
Several noise-robust losses have been proposed for training models with noisy labels \cite{reed2014training,zhang2018generalized,wang2019symmetric,ma2020normalized,menon2020can,jin2021instance,zhou2021learning}, which were shown to be more robust than CE. 
However, since current deep networks have a large number of parameters, these methods can still memorize the noisy labels given sufficient training time \cite{zhang2017understanding}. 

Another direction is to identify noisy data or clean data and cope with them separately either by re-weighting them or converting the problem to a semi-supervised learning task.
\cite{arpit2017closer,charoenphakdee2019symmetric} found out the memorization effect which is stated as although deep networks can memorize noise data, they tend to learn simple patterns first. 
Based on the memorization effect \cite{arpit2017closer,zhang2021understanding}, many methods separate clean and noisy samples by using loss value \cite{han2018co,jiang2018mentornet,arazo2019unsupervised,shu2019meta,yao2019searching,li2020dividemix} or forgetting events \cite{malach2017decoupling,yu2019does}. 
%The re-weighting methods \cite{ren2018learning,shu2019meta} learned optimal weights for different samples by using meta-learning \cite{hospedales2020meta}. 
%The semi-supervised learning methods \cite{kim2019nlnl,huang2019o2u,li2020dividemix} divide the training data into a labeled set with clean samples and an unlabeled set with noisy samples, and trains the model on both the labeled and unlabeled data in a semi-supervised manner.
The setting of traditional noisy labels learning does not consider the existence of a pre-trained model. 
\section{Method}
Due to the versatility of LaVeEntail \cite{sainz2021label, saintz2022textual} across multiple tasks such as RE and EAC, we employ LaVeEntail as the backbone to obtain silver standard data.
We will introduce the LaVeEntail method in \S \ref{sec:LaVeEntail}; 
the clean data detection module of Clean-LaVe in \S \ref{sec:clean_label_detect}; 
the finetuning and inference stage in Clean-LaVe in \S \ref{sec:finetune_infer}. 
%, including weighted negative learning in \S \ref{sec:weight-NL} and class-aware data selector in \S \ref{sec:data selector};
%the finetuning and inference stage in Clean-LaVe in \S \ref{sec:finetune_infer}. 
% \revisezq{check it. it is old. no mention the new name: weighted negative learning and class-aware data selector.}
% \revisejw{they are mentioned in line 7-8 of this paragraph.}

\subsection{LaVeEntail}
\label{sec:LaVeEntail}
LaVeEntail \cite{sainz2021label,saintz2022textual} includes two processes for relation extraction and event argument extraction, i.e., label verbalization and textual entailment model inference. 
% We further extend the capabilities of LaVeEntail to encompass other information extraction, such as event extraction and event argument extraction.

\subsubsection{Label Verbalization}
The label verbalization process creates templates of classes (i.e., relation types and argument roles) and then uses them to generate hypotheses. The templates can be easily created because relation labels and argument roles naturally implicate such verbalization templates. 
For example, the relation \texttt{per:schools\_attended} can be verbalized as \texttt{\{subj\} studied in \{obj\}}, where \texttt{\{subj\}} and \texttt{\{obj\}} are placeholders for subject and objective entities. 
For example, \texttt{giver} can be expressed as \texttt{\{arg\} gave something to someone}, where \texttt{\{arg\}} is the placeholder for an argument span. 
%The role corresponding to the template with the highest score is assigned to the argument span. 
%Given an input sentence $x$ with subject entity being $x_{subj}$ and objective entity being $x_{obj}$, the hypothesis is generated by substituting  $x_{subj}$ and  $x_{obj}$ to corresponding placeholders, i.e., \texttt{\{subj\}} and \texttt{\{obj\}}. 

\subsubsection{Textual Entailment Model Inference}
For each input sentence, LaVeEntail constructed hypotheses that are generated by verbalization templates of all relation types (or argument roles), and fed them to a TE model, and obtained entailment scores of all hypotheses. 
LaVeEntail inferred that the predicted relation (or role) type of the input sentence is the relation (or role) type whose hypothesis yields the highest entailment score. 
Figure \ref{fig:procedure} shows the inference procedure of relation extraction.

Entity type information is helpful to infer relation types \cite{tran2020revisiting}. 
A relation naturally indicates entity types of subject and object. 
For instance, the relation \texttt{per:city\_of\_death} implicates that the entity type of subject and object should be \texttt{PERSON} and \texttt{CITY} respectively.
In the inference stage, when the entity type information is given, we could rule out some relation types that are impossible to be ground truth.
LaVeEntail created entity type constraint(s) for each relation according to the meaning of the relation.
If the entity types in the input sentence do not match the entity type constraints of a relation, then the entailment score(s) of all hypotheses related to this relation is set to zero.  In the case where there is no relation between two entities, a threshold-based approach is used to detect \texttt{no\_relation}. 
If the entailment scores of all hypotheses are less than a threshold, the prediction is \texttt{no\_relation}. 

% \subsubsection{Extension to Other Information Tasks}
% Since LaVeEntial doesn't rely on task-related information, this approach of transforming relation extraction task into text entailment can be seamlessly applied to  other information extraction tasks with rewritten verbalization.

% \textbf{Event Extraction (EE)}\cite{huang2017zero,lyu2021zero,lin2020joint} seeks to detect events occurring in the sentence. Each event type can be verbalized as a template, e.g. \texttt{meet} is represented by \texttt{It mentions face-to-face talks}. TE model is employed to obtain confidence for event templates and events with confidence exceeding the predefined threshold are assumed to have occurred.   

% \textbf{Event Argument Classification (EAC)} \cite{huang2017zero,lyu2021zero,lin2020joint} aims to categorizing each event argument into appropriate role. Following by prior verbalization step, each role can be represented by a template, for example, \texttt{giver} can be expressed as \texttt{\{arg\} gave something to someone.} \texttt{\{arg\}} is the placeholder for argument entity. The role corresponding to the template with the highest score is assigned to the argument. 

\subsection{Clean Data Detection}
\label{sec:clean_label_detect}
The clean data detection module aims to select relatively clean data for subsequent finetuning from silver standard data annotated by LaVeEntail.
To alleviate the impact of imbalanced noisy data, we introduce an iteratively weighted negative learning (IWNL) algorithm. Additionally, we employ a class-aware data selector (CADS) to choose clean samples from a boarder range of classes.
%Clean data detection module leverages the silver data generated by LaVeEntial method and selects clean data for subsequent finetuning process. 
%\revisezq{The silver data is annotated with confidence scores using the Weighted Neagtive Learning, which is built upon Negative Learning for Noisy Labels (NLNL) \cite{kim2019nlnl}.} 
%\revisezq{Additionally, the Class-aware Data Selector is employed to select data from a broader range of classes.}

\subsubsection{Iteratively Weighted Negative Learning}
\label{sec:weight-NL}
Negative Learning \cite{kim2019nlnl} loss is robust to noise. 
Different from positive learning loss (e.g., cross entropy loss) which tells the model what is correct, the negative learning loss provides the model with the complementary label(s), telling what is not correct, e.g., the input image is not a dog. 
The complementary label is randomly selected from the label space excluding the input label (possibly noisy). 
For noisy data, the probability of selecting the ground truth as the complementary label is low. 
Hence, using negative learning loss can decrease the risk of overfitting noisy labels. The formula of NL loss is shown as follows,
\begin{equation}
    \label{eq:loss_func}
     \mathcal{L}_{neg}=- \sum_{d \in D} \sum_{i=1}^{|\YM|} \widehat{\y}_{i}^{d} \log (1-\p_{i}^{d}),
\end{equation}
where $d$ is a sample in the dataset $D$, $|\YM|$ is the number of relation types, $\widehat{\y}^{d}$ is a one-hot vector with the complementary label being one, $\widehat{\y}_{i}^{d}$ is the $i$-th element of $\widehat{\y}^{d}$, $\widehat{\p}^{d}$ is the output probability distribution of a smaple $d$, and $\widehat{\p}_{i}^{d}$ is the $i$-th element of $\widehat{\p}^{d}$.

The original NL loss in equation (\ref{eq:loss_func}) treats each class equally, which may not be appropriate when dealing with real-world datasets that exhibit severe class imbalance. 
In these datasets, majority classes often have a significantly higher number of data samples compared to minority classes. 
Consequently, the model encounters much fewer samples in the minority classes, leading to underfitting (i.e., high loss values) during the training process. 
It poses a challenge to distinguish between clean and noisy samples in minority classes as they both have high loss values. 
%Consequently, applying the original NL loss to an imbalanced dataset would result in ignoring the minority classes. 
We propose a iteratively weighted NL loss to alleviate this issue, giving more weight on minority classes.
\begin{eqnarray}
    \label{eq:wloss_func1}
     \mathcal{L}_{neg}^{j}&=&- \sum_{d \in D} \sum_{i=1}^{|\YM|} w_{i}^{j} \cdot \widehat{\y}_{i}^{d} \log (1-\p_{i}^{d}) \\
     \label{eq:wloss_func3}
     w_{i}^{j}&=&w_{i}^{j-1} \cdot e^{1 - \frac{c_i^{j-1}}{c_\AM^{j-1}}} \\
     \label{eq:wloss_func2}
     w_{i}^{0}&=&\frac{\sum_{k=1}^{|\YM|}c_k^{0}}{c_i^{0}} 
\end{eqnarray}
where $\mathcal{L}_{neg}^{j}$ represents the negative loss for $j$-th epoch, $w_{i}^{j}$ denotes the weight for class $i$ in $j$-th epoch, which is dynamically updated by prediction in previous epoch according to equation (\ref{eq:wloss_func3}). 
${c_i^{j-1}}$ is the quantity of class $i$ in $j-1$ th epoch and ${c_\AM^{j-1}}$ is the average quantity across all classes in $j-1$ th epoch. The quantity of class $i$ is the number of samples that are predicted as class $i$. 
Initial weight $w_{i}^{0}$ is computed according to labels of silver data, as described in equation (\ref{eq:wloss_func2}). 
According to Eq. (\ref{eq:wloss_func2}) and Eq. (\ref{eq:wloss_func2}), minority classes have more weight. 
If the dataset is initially balanced, our IWNL algorithm can degenerate to the original negative learning algorithm. 

We use BERT \cite{devlin2018bert} as the relation classifier when using  IWNL loss.
Although it is possible to train a TE model as a relation classifier using IWNL loss, its performance falls short of that of BERT-based classifier. 
%We provide a detail comparison in Experiment \ref{exp:classifier}. 
After being trained with IWNL loss, the classifier attempts to assign high confidence scores to clean data while give low confidence scores to noisy data. 
These confidence scores can be leveraged for subsequent data selection.

\subsubsection{Class-Aware Data Selector}
\label{sec:data selector}
A straightforward approach to selecting clean data is sorting all samples according to their confidence scores and then selecting a fixed proportion $\eta$ of whole data as the clean data set. Given that $\SM(D_s)$ is the total confidence scores of all samples in $D_s$, and $\eta$ is a hyperparameter representing selection proportion, the clean data set $D_{clean}$ are selected as follows, 
\begin{equation}
\label{eq:select_NL}
    {D}_{clean} = {\arg\max}_{D_s:\mid D_s \mid = \eta \cdot \mid D_{silver} \mid} \SM(D_s).
\end{equation}
However, it does not consider class diversity. 
Samples in some classes can yield very high confidence scores while some classes have very low confidence scores. 
Large quantities of samples in those classes are selected in $D_{clean}$ while some classes even do not have any clean data in $D_{clean}$, which harms performance badly. 

We propose a class-aware data selection algorithm that considers confidence scores as well as class diversity.
First, we select a proportion $\eta$ of data with high confidence scores. 
This step can ensure that samples with low noise levels are selected. 
Next, we select $m$ more samples to encourage diversity. 
For each class, we select some samples with high confidence scores in this class. 
The number of selected samples for a class is proportional to the number of samples that are predicted as the class.
%Ideally, we should select samples according to the true quantity of each class. 
%The quantity of a class is the number of samples that are labeled as the class.
%But it is unavailable, we estimate it based on the quantities observed in silver prediction.  
%This step aims to involve samples from more classes. 
The class-aware data selection algorithm is presented in Algorithm \ref{alg:class_aware}.
% The number of selected samples of a class is dynamic, which is determined by the number of samples of a class in the silver candidate pool. The counterpart of the dynamic strategy is selecting a fixed number of samples for each class. We compare dynamic and fixed methods in the experiment. \revisezq{Do we still have dynamic and fixed?}

The low confidence scores observed in certain classes can be attributed to two factors: either they are minority classes and suffer from underfitting, or the samples in these classes are noisy. 
The class-Aware data selector serves as a compensatory mechanism to mitigate the impact of underfitting in minority classes.
However, class-aware data selectors carry the risk of inadvertently noisy data. This risk becomes even more pronounced when the dataset is balanced, as evidenced by the experimental results in Table \ref{tab:main_result}.

\begin{algorithm}[tp] 
    \caption{Class-Aware Data Selector}
    \label{alg:class_aware}
    \begin{flushleft}
        \textbf{Input:} silver standard data set $D_{silver}$, proportion $\eta$, diversity number $m$, the set of classes $\CM$, the total confidence scores function $\SM(\cdot)$.
    \end{flushleft}
    \begin{algorithmic}[1]
        \STATE $D_{clean} = \varnothing $.
        \STATE Obtain  $D_{clean}$ using Eq. \ref{eq:select_NL} by setting the proportion to $\eta$.
        \STATE $D_{rest}= D_{silver} - D_{clean}$, divide  $D_{rest}$ into $|\CM|$ subsets according to class predictions. The subset for class $c$ is denoted as $D^{c}$. 
        \FOR{ $c\;$ \text{in} $\;\CM$ }
            \STATE ${D}_{clean}^{c} = {\arg\max}_{D_s:\mid D_s \mid = \frac{\mid D^{c} \mid}{\mid D_{rest}\mid } \cdot m }          \SM(D_s)$
            \STATE $D_{clean} = D_{clean} \cup D^{c}_{clean}$
        \ENDFOR
    \end{algorithmic}
    \begin{flushleft}
        \textbf{Output:} clean data set $D_{clean}$.
    \end{flushleft}
\end{algorithm}

\subsection{Finetuning and Inference}
\label{sec:finetune_infer}
After running clean data detection algorithms, we obtain $D_{clean}$ which consists of the input sentence and its relation (role) type pairs. 
Since the input formats of the TE and RE (or EAC) tasks are different, we need to convert $D_{clean}$ to premise-hypothesis pairs so that we can use $D_{clean}$ to finetune the TE model.

For each relation (or role) in the RE (or EAC) task form, we should create entailment, contradiction, and neutral hypothesis for the TE task. 
The entailment hypothesis is generated with the templates that describe the
ground truth relation (or role), the neutral hypothesis is generated by randomly select a template that does not describe the ground truth relation (role) and the contradiction hypothesis is generated using the template ``\texttt{\{subj\}} and \texttt{\{obj\}} are not related,'' or ``\texttt{\{arg\}} '' is not an argument of \texttt{\{trg\}} where \texttt{\{trg\}} is the trigger word.

We use premise hypothesis pairs constructed from $D_{clean}$ to finetune the off-the-shelf TE model. Finally, we use the finetuned TE model to infer relation (role) types on the test set. The complete algorithm is presented in Algorithm \ref{alg:whole}. 

\begin{algorithm}[tp] 
    \caption{Finetuning and Inference}
    % \caption{Zero-shot Relation Extraction with Clean Data Finetuning}
    \label{alg:whole}
    \begin{flushleft}
        \textbf{Input:} silver standard data set $D_{silver}$, test set $D_{test}$, textual entailment model $\MM$.
        
    \end{flushleft}
    \begin{algorithmic}[1]
        % \STATE Verbalize relation label
        % \STATE Generate hypothesis of $D$
        % \STATE Use $NLI$ infer pseudo label set $D_{silver}$
        \STATE Obtain  $D_{clean}$ using Algorithm \ref{alg:class_aware}. 
        \STATE Generate premise hypothesis pairs dataset  $D_{clean}^{'}$ based on $D_{clean}$.
        \STATE {Finetune $\MM$ using $D_{clean}^{'}$, and obtain finetuned model $\MM'$.}
        \STATE Use $\MM'$ to infer relation (role)types on $D_{test}$.
    \end{algorithmic}
    \begin{flushleft}
        \textbf{Output:} relation (role) types of samples on $D_{test}$.
    \end{flushleft}
    
\end{algorithm}

\begin{table}[H]
    \centering
    \resizebox{0.5\textwidth}{!}{
        \begin{tabular}[H]{ccccccc}
            \hline

          \multirow{2}{*}{ Dataset} & \multirow{2}*{Relation Types} & \multirow{2}*{Entity Types} & \multirow{2}*{Distribution} & \multicolumn{3}{c}{Instances} \\ &&&&Train & Dev & Test \\
          
          \hline
          
          TACRED & 42 & 17 & Skewed & 68124 & 22631 & 15509 \\
          Wiki80 & 80 & 29 & Uniform & 40320 & 10080 & 5600 \\ 
          Smiler-It & 22 & - & Skewed & 73228 & 746 & 1510 \\ 
          Smiler-Po & 21 & - & Skewed & 16651 & 180 & 344 \\ 
          Smiler-Kr & 28 & - & Skewed & 18538 & 173 & 382 \\ 
         ACE05-E+ & 22 & 7 & Skewed & 4815 & 603  & 573  \\ 
          \hline
        \end{tabular}
    }
    \caption{
    The statistics of datasets. 
%    Each instance is a sentence with two entities and their entity types.
    }
    \label{tab:statistics}
\end{table}
\section{Experiment}
\subsection{Experiemental Settings}
\label{sec:data_statistics}
We follow the same zero-shot setting with LaVeEntail, all data used for training are unlabeled, and only 1\% development set are available for adjusting hyperparameters. 
For the zero-shot cross-lingual setting, we do not use any annotated training data from both the source language and target language. 
Traditionally, zero-shot cross-lingual methods \cite{huang2022multilingual, ma2023shine, ma2022wider, wu2020single} used a large number of annotated data from the source language. 
%only 2 examples per relation (1\% dev set)
%We follow the same zero-shot setting with LaVeEntail, all data used for training are unlabeled and only 2 examples per relation (1\% dev set) are available only for adjusting hyperparameters.

For the zero-shot RE task, we evaluate our method on the TACRED \citeplanguageresource{TacRE} and Wiki80 \citeplanguageresource{Wiki80} dataset. 
For the zero-shot cross-lingual RE task, we evaluate our method on the Smile \citeplanguageresource{Smiler} dataset which contains 14 languages.
We evaluate Clean-LaVe in three languages, i.e., Italian, Polish, and Korean.  
For the zero-shot EAC task, we evaluate our method on ACE05-E+ \citeplanguageresource{lin2020joint}.
The statistics of datasets are shown in Table \ref{tab:statistics}.
%Additionally, we utilize WikiFact \citeplanguageresource{Wikifact} as extra silver data. We don't access the labels of data but use TE to annotated them. 

The TE model we used for RE and EAC tasks is microsoft/deberta-v2-xlarge-mnil \cite{he2021deberta}.
We use mDeBERTa-v3-base-xnli-multilingual-nli-2mil7 \cite{laurer_less_2022} for the cross-lingual RE task as it can process multiple languages.
%a large-scale relation extraction dataset encompassing 923 relation types.

For each dataset, we manually created verbalization templates for each relation or argument role, as well as entity type constraints.
The constraints we used on TACRED dataset are different from those of LaVeEntail. We delete the constraints which leak the information of the ground truth. 
For example, there is only one relation type that has the constraint where the subject entity type is \texttt{PERSON} and the object entity type is \texttt{TITLE}. 
The sentence that satisfies this constraint has a very large probability of being inferred as \texttt{per:title} relation because other relations are ruled out. 
\citep{tran2020revisiting} showed that entity types are a strong inductive bias. However, in LaVeEntail, the inductive bias is not learned by the algorithm itself but by manually designed type constraints. It leads to artificially inflated performance, so we deleted those type constraints.
Besides, we also conduct experiments using original type constraints in \S \ref{exp:full_constraint}.

\begin{table*}[!t]
	\centering
	%\scalebox{0.98}{
	\scalebox{0.78}{
		\begin{tabular}{l |  p{1.75cm}<{\centering}  p{1.75cm}<{\centering} | p{1.75cm}<{\centering}   p{1.75cm}<{\centering}   p{1.75cm}<{\centering}  | p{1.75cm}<{\centering} }
			\toprule
			 & \multicolumn{2}{c|}{RE} & \multicolumn{3}{c|}{Cross-lingual RE} & EAC \\	
                \cmidrule(lr){2-7}
	        % 左边是TACRED, 右边是Wiki80
			&   TACRED & Wiki80 & Smiler-It & Smiler-Po & Smiler-Kr & ACE05-E+  \\
			\midrule 
			CE 
			& 45.35\small{$\pm 0.58$} & 40.76\small{$\pm 0.29$}		& 40.79\small{$\pm 0.12$} & 41.56\small{$\pm 0.19$} & 49.75\small{$\pm 0.50$} & 71.79\small{$\pm 0.96$}  \\	
			GCE \cite{zhang2018generalized}
			&45.93\small{$\pm 0.67$} &41.28\small{$\pm 0.61$}		& 47.27\small{$\pm 0.21$}  & \uline{45.99\small{$\pm 0.60$}} & 53.35\small{$\pm 0.49$} & 71.61\small{$\pm 0.79$} \\	
			SCE \cite{wang2019symmetric}
			&45.82\small{$\pm 0.92$} &41.12\small{$\pm 0.24$} &  40.97\small{$\pm 0.70$} & 40.41\small{$\pm 0.31$} & 47.79\small{$\pm 0.09$} & 71.88\small{$\pm 0.26$} \\	
			Co-Regularization  \cite{zhou2021learning}
			& 48.86\small{$\pm 0.34$}& 28.48\small{$\pm 0.42$}		&  42.17\small{$\pm 0.61$}& 41.86\small{$\pm 0.19$}  & 50.16\small{$\pm 0.70$} & \uline{72.93\small{$\pm 0.17$}}\\
		    \midrule 
			O2U  \cite{huang2019o2u}
			& 47.52\small{$\pm 0.81$} & 42.62\small{$\pm 0.03$}	& 41.12\small{$\pm 0.23$} & 44.47\small{$\pm 0.66$} & 49.67\small{$\pm 0.49$} & 69.83\small{$\pm 0.06$}\\	
			DivideMix  \cite{li2020dividemix}
			& 49.78\small{$\pm 0.80$} & \uline{45.52\small{$\pm 0.26$}} 		& 41.94\small{$\pm 0.78$} & 43.79\small{$\pm 0.69$}  &  52.48\small{$\pm 0.80$} & 69.13\small{$\pm 0.64$} \\
			\midrule
            Global\_Constraints \cite{lin2023constraintsEAC}
               - &  - & -  & - & - & - & \multicolumn{1}{c}{\text{66.1$^{*}$}} \\
                QA4RE \cite{zhang2023QA4RE}
                &  \uline{58.55\small{$\pm 0.05$}} & 43.93\small{$\pm 0.09$}  		&  \uline{\textbf{56.42}\small{$\pm 0.84$}} & 38.09\small{$\pm 0.19$}  & \uline{56.08\small{$\pm 0.73$}} & 64.74\small{$\pm 0.84$}\\
            \midrule    
            LaVeEntail\protect\footnotemark[1] \cite{sainz2021label}
            & 52.18 & 41.16	& 39.96 & 37.84  & 44.30  & 	71.60 \\
            %\multicolumn{7}{l}{\emph{$\rightarrow$ Classifiers based noise-robust learning} } \\
			Labeled Data Finetune (1\%)
			& 56.61\small{$\pm 1.29$} & 47.39\small{$\pm 0.33$} 	& 51.85\small{$\pm 0.96$} & 46.44\small{$\pm 0.66$} & 47.33\small{$\pm 0.91$}& 	76.21\small{$\pm 1.50$} \\
			Labeled Data Finetune (5\%)
			& 63.72\small{$\pm 1.03$} & 53.89\small{$\pm 0.46$} 	& 52.56\small{$\pm 0.57$} & 49.56\small{$\pm 0.54$} & 55.30\small{$\pm 0.14$}&  78.87\small{$\pm 0.17$}	  \\
            \midrule
			\textbf{Silver-LaVe}
			&54.67\small{$\pm 0.58$} & 44.57\small{$\pm 0.31$}& 48.91\small{$\pm 0.55$} & 50.60\small{$\pm 0.38$} & 54.64\small{$\pm 0.81$}  & 	80.18\small{$\pm 0.08$}	\\
               
                % \textbf{Clean-LaVe}
                % & \textcolor{red}{\textbf{63.36\small{$\pm 1.03$}}} $\uparrow5\%$ & 
                %  51.53\small{$\pm 0.53$} $\downarrow1\%$ & 81.22\small{$\pm 0.38$} $\downarrow1\%$	& 55.09 $\downarrow1\%$	& 52.99\small{$\pm 0.88$} $\downarrow12\%$   & \textcolor{red}{\textbf{59.41\small{$\pm 0.84$}}} $\uparrow3\%$
                % \\
                 \textbf{Clean-LaVe}
                & \textbf{63.36\small{$\pm 1.03$}}  & 
                 51.53\small{$\pm 0.53$}   	& 55.09\small{$\pm 0.05$} 	& \textbf{52.99}\small{$\pm 0.88$}   & \textbf{59.41\small{$\pm 0.84$}} & \textbf{81.22}\small{$\pm 0.38$}
                \\

                 -- Iteratively Weighted Negative Learning 
                &58.66\small{$\pm 0.93$}& 48.44\small{$\pm 0.44$}	& 54.20\small{$\pm 0.97$}	& 48.09\small{$\pm 0.59$} &  57.18\small{$\pm 0.95$}& 78.07\small{$\pm 0.82$}
                \\
                 -- Class-Aware Data Selector 
                & 59.55\small{$\pm 0.98$} & \textbf{52.52\small{$\pm 0.21$}} 	& 54.97\small{$\pm 0.33$} & 50.14\small{$\pm 0.64$}  & 57.34\small{$\pm 0.26$}& 	78.14\small{$\pm 0.61$}
                \\
                -- Above Both
                & 56.41\small{$\pm 1.82$} & 52.34\small{$\pm 0.38$} 	& 54.28\small{$\pm 0.65$} & 45.99\small{$\pm 0.41$} & 54.97\small{$\pm 0.74$} & 	77.37\small{$\pm 0.67$}
                \\
                %\midrule
                %\multicolumn{7}{l}{\emph{$\rightarrow$ Methods based on Chatgpt} } \\
			\bottomrule
    	\end{tabular}}
	\caption{Results of zero-shot classification tasks. We report the average of micro F1 scores in 3 runs. 
The best F1 scores are marked in \textbf{bold}. 
SOTA baselines are highlighted with \uline{underline}.
Results marked with * are retrieved from the original paper.}
    \label{tab:main_result}
\end{table*} 
\footnotetext[1]{LaVeEntail direct infers on the test set and does not involve any training process, resulting in zero variance.}

\subsection{Compared Methods}
To demonstrate the effectiveness of our method, we compare our model with the following baselines: 

We consider training a supervised relation classification model on silver standard data using \textbf{CE} (Cross Entropy loss) and different noise-robust losses including \textbf{GCE} (Generalized Cross Entropy loss) \cite{zhang2018generalized}, \textbf{SCE} (Symmetric Cross Entropy loss) \cite{wang2019symmetric}, and \textbf{Co-Regularization} \cite{zhou2021learning} as baselines. 
%\textbf{BSH} (Bootstrap Hard loss) \cite{reed2014training}, 
%\textbf{ER-GCE} (Entropy Regularized Generalized Cross Entropy loss) \cite{jin2021instance}

We also include the following representative noise-robust learning algorithms which identify noisy data or clean data and convert the problem to a semi-supervised learning problem as baseline methods: 
\textbf{O2U} (Overfiting to Underfitting) \cite{huang2019o2u}, and \textbf{DivideMix} \cite{li2020dividemix}.

We also consider the SOTA zero-shot RE method \textbf{QA4RE} \cite{zhang2023QA4RE} and the SOTA zero-shot EAC method \textbf{Global\_Constraints} \cite{lin2023constraintsEAC} as baselines. 
QA4RE is based on ChatGPT \cite{gpt4}, which is easily adapted to the zero-shot EAC task.
Global\_Constraints is delicately designed for the zero-shot EAC task. 
%\textbf{NLNL} (Negative Learning for Noisy Labels),
%\textbf{Self-training} \cite{yarowsky1995unsupervised} first trained a classifier on a small amount of labeled data in a supervised manner, and then used this classifier to annotate more samples to train this classifier again. 
%The initial labeled data are annotated by an off-the-shelf TE model. 
%Only samples with a confidence score greater than a threshold are selected into the initial labeled data set. 
%The threshold is selected on the development set. 

%\textbf{O2U} (Overfiting to Underfitting) \cite{huang2019o2u} changed the model status from underfitting to overfitting repeatedly, and then used the loss to detect and remove noisy data, and finally trained the model using clean data. We directly apply it to silver standard data.
%\textbf{NLNL} (Negative Learning for Noisy Labels) \cite{kim2019nlnl} first trained a classifier by using NL loss, then and selected partial data and trained them using NL as well as positive learning loss, and finally selected some clean data as labeled data and trained a classifier in a semi-supervised manner.  We directly apply it to silver standard data. 
%\textbf{DivideMix} \cite{li2020dividemix} used a Gaussian Mixture Model (GMM) to divide the training data into a labeled set with clean samples and an unlabeled set with noisy samples, and trained the classifier on both the labeled and unlabeled data in a semi-supervised manner.  
%It is a competitive method in noise labels learning area. 
%We directly apply it to silver standard data.

\textbf{LaVeEntail} \cite{sainz2021label} utilized an off-the-shelf textual entailment model to directly infer the test data.
\textbf{Labeled Data Finetune} randomly select a proportion of labeled training data to finetune the off-the-shelf TE model. 
\textbf{Clean-LaVe} is our proposed method. Additionally, we conduct comparisons by removing the weighted negative learning module, the class-aware data selector, and both of them respectively to assess their impact on the results. 
\textbf{Silver-LaVe} can be considered as Clean-LaVe without a clean data detection module, which uses all silver standard data to finetune an off-the-shelf TE model. 
%\textbf{Clean-LaVe + Extra Silver Data} leverages extra silver data from external corpora to achieve further performance improvement.

\subsection{Result Analysis}
%In the first block of Table \ref{tab:main_result}, we show results of LaVeEntail 
%and results by fine-tuning a Textual Entailment (TE) model using 1\% and 5\% labeled training data. 
%Clean-LaVe is comparable to LaVe using 5\% labeled training data. 
As shown in the first and second block of Table \ref{tab:main_result}, Clean-LaVe outperforms noise-robust loss based methods and semi-supervised based noisy labels learning methods across all datasets. 
Directly applying noisy labels learning methods on silver standard data is straightforward but not effective. 
Hence, there is a need to investigate how to use silver data.

As shown in the third block, Clean-LaVe outperforms the SOTA methods by 3\% \textasciitilde 15\% on all datasets except on the Smiler-It. 
On Smiler-It, QA4RE outperforms Clean-LaVe by 1\%. 
Despite facing a stronger competitor based on ChatGPT, Clean-LaVe delivers commendable overall performance.
%The possible barriers to good performance are training a classifier from scratch and the high noise ratio of training data. 
%As shown in the first block of Table \ref{tab:main_result}, noise-robust loss based methods cannot outperform LaVeEntail in TACRED and are comparable with LaVeEntail in Wiki80. 
%The possible barriers to good performance are training a classifier from scratch and the high noise ratio of training data. 
%In the second block, the semi-supervised based noisy labels learning methods are better than noise-robust loss based method. The best performance is DivideMix.

As shown in the fourth block, Clean-LaVe can gain significant improvement compared to LaVeEntail by 10\% \textasciitilde 16\%. 
Additionally, our method is comparable to or even outperforms the supervised LaVeEntail with 5\% labeled data.

As shown in the last block, we surprisingly find Silver-LaVe outperforms LaVeEntail by 2\% \textasciitilde 13\%. It indicates that, to some content, our proposed framework (i.e., finetuning pre-trained model with silver standard data) can be beneficial, regardless of the quality of silver standard data. Clean-LaVe outperforms Silver-LaVe, indicating the effectiveness of the clean data detection module.

%Additionally, it inspires us to explore ways to utilize silver data more effectively for finetuning pre-trained models.

We also provide results after removing Iteratively Weighted Negative Learning, Class-Aware Data Selector, and both of them respectively. 
After removing the IWNL component, we observe decreases in performance across all datasets, which validates the effectiveness of this component. 
After removing the CADS, we observe decreases in performance across all datasets except Wiki80, which validates the effectiveness of this component. 
Removing the CADS leads to a slight improvement (1\%) on Wiki80. 
This improvement can be attributed to the fact that Wiki80 is a balanced dataset. 
The reason has been stated in \S \ref{sec:data selector}.

\subsection{Case Analysis}
\begin{figure}[t]
	\centering
	\subfigure[Analysis of IWNL on minority and majority classes.
% Case study of iteratively weighted negative learning (IWNL). Clean data detection accuracy scores under different proportion $\eta$ and accuracy scores of selected classes under the best proportion $\eta$.
 ]{\label{fig:vis1}
		\includegraphics[width=0.48\textwidth]{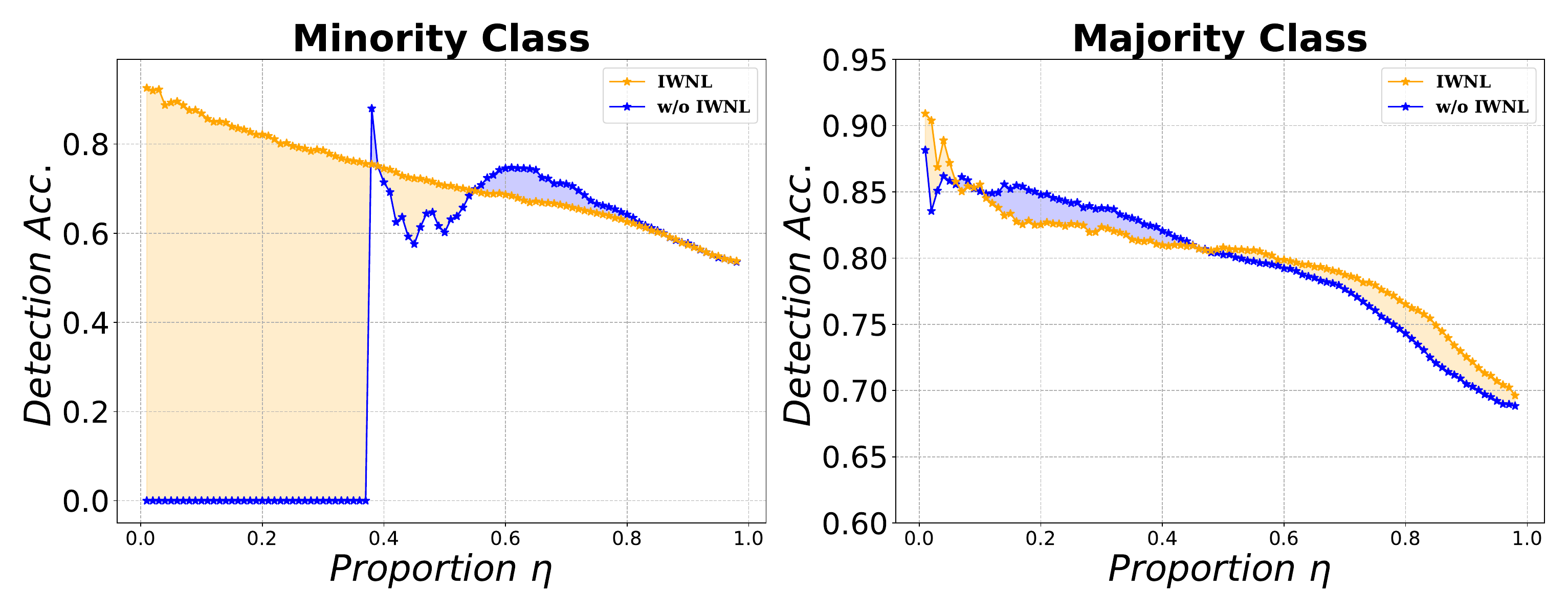}
	}
	\subfigure[
 Analysis of CADS.
 %Case study of class-aware data selector (CADS). Clean data detection accuracy scores of minority classes and the number of samples of selected classes.
 ]{\label{fig:vis2}
		\includegraphics[width=0.48\textwidth]{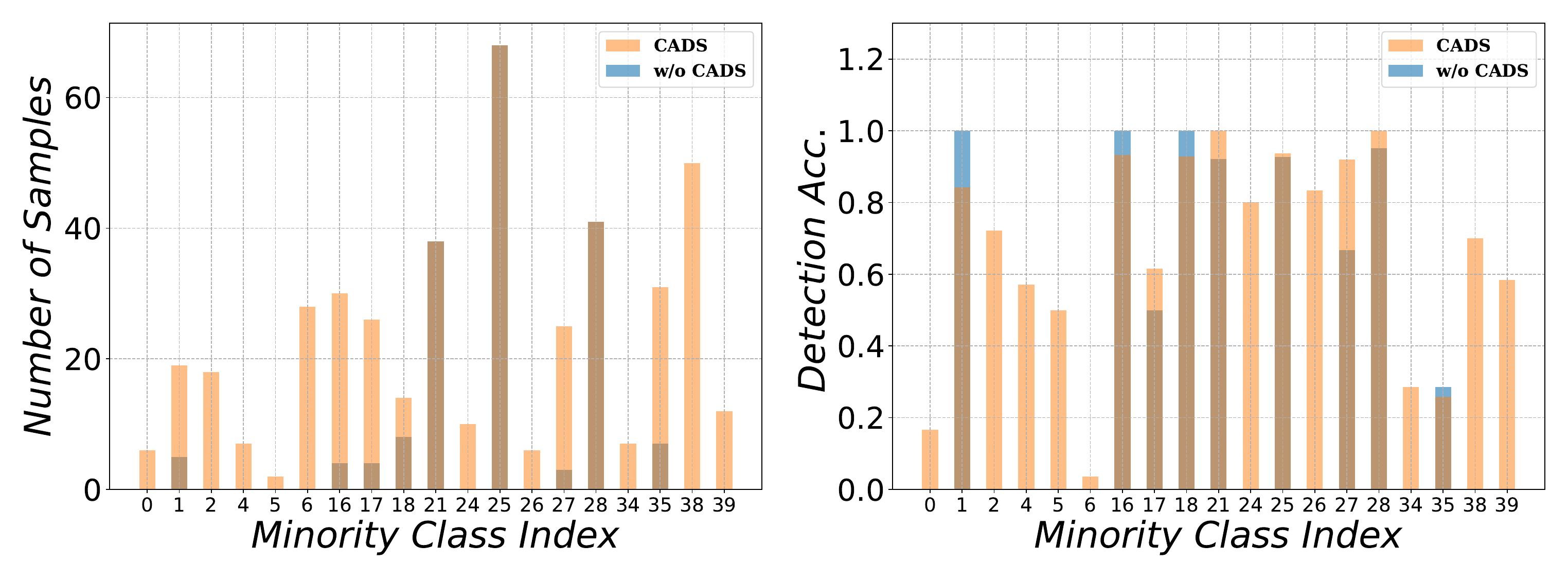}
	}
	\vspace{-0.1in}
	\caption{Analysis of IWNL and CADS on TACRED.
	}	
 \label{fig:case_study}
\end{figure}

%We conduct an in-depth analysis regarding the effectiveness of Iteratively Weighted Negative Learning (IWNL) and Class-Aware Data Selector (CADS) in the Clean Data Detection module. 
We conduct an in-depth analysis on TACRED regarding the effectiveness of Iteratively Weighted Negative Learning (IWNL) and Class-Aware Data Selector (CADS). 
%The case study is conducted for the zero-shot RE task on TACRED.
The clean data detection accuracy (referred to as detection accuracy for brevity) is the percentage of clean data whose predictions are equal to ground truths. 
We sort the class according to the number of samples in the class in descending order and consider the former (latter) half as the majority (minority) classes.
The proportion $\eta$ indicates how much proportion of data is selected as clean data.

\textbf{Iteratively Weighted Negativing Learning (IWNL)} can alleviate the effect of underfitting and improve the clean data detection accuracy of minority classes. 
As depicted in Figure \ref{fig:vis1} (left), IWNL yields consistently higher detection accuracy scores than w/o IWNL on minority classes. 
Without IWNL, the detection accuracy for minority classes is almost negligible when the selection proportion is small.
As depicted in Figure \ref{fig:vis1} (right), the performance of IWNL on majority classes is comparable to w/o IWNL.  
%\revisejw{
%After applying IWNL, the detection accuracy of minority classes increase, suggesting IWNL can indeed alleviate the effect of underfitting. 
%Iteratively Weighted Negativing Learning (IWNL) performs better than w/o IWNL under low proportions and falls behind with proportion increasing shown in the left figure of Figure \ref{fig:vis1}. 
%In low-proportion scenarios, both methods focus solely on the detection of major classes, with IWNL contributing to a cleaner detection process for these major classes. 
%However, as the proportion increases, minor classes may also be included.
%IWNL encourages a higher level of attention toward the minor classes and incorporates more samples from the minority class, as depicted in the right figure of Figure \ref{fig:vis1}.
%Although the overall accuracy is slightly lower compared to the method without IWNL, IWNL detects more samples from minority classes with considerable accuracy.
%}

\textbf{Class-Aware Data Selector (CADS)} can encourage clean samples from a broader range of classes. 
As depicted in Figure \ref{fig:vis2} (left), there are more orange bars than blue bars, indicating CADS selects samples from more classes, especially from minority classes. 
As depicted in Figure \ref{fig:vis2} (right), the detection accuracy scores of classes that are only selected by CADS are satisfying overall. 
For classes that are selected by both CADS and w/o CADS, the accuracy scores of some classes increase but some decrease after applying CADS. 
The possible reason for decreased accuracy is that it involves noisy data, as we have discussed in \S \ref{sec:data selector}. 

\subsection{Full Constraints Comparison}
\label{exp:full_constraint}
As previously mentioned, we remove some constraints defined in LaVeEntail to prevent information leakage.
In this section, we compare our method with LaVeEntail using full constraints. 
%In this section, we compare our method with LaVeEntail and LaVeEntail-BIRD using full constraints. 
%LaVeEntail-BIRD \cite{rahimi2023improving} automatically generate hypothesis templates given manual templates as seeds. 
As table \ref{tab:full_constraints} is shown, our method still outperforms LaVeEntail given full constraints. 
%We believe that our methods can also benefit from BIRD \cite{rahimi2023BIRD} technique and achieve better peformance. 
%the automatically generated templates in LaVeEntail-BIRD are not public available. 
Under full type constraints, LaVeEntail and Clean-LaVe obtain improvement compared to partial constraints results in table \ref{tab:main_result}. But the improvement is inflated.  

\begin{table}[htbp]
    \centering
    \renewcommand\arraystretch{1.5}
    \resizebox{0.45\textwidth}{!}{
        \begin{tabular}[H]{c c | p{1cm}<{\centering} | p{1cm}<{\centering}  | p{2cm}<{\centering}}
             \midrule
           &  & Pr. & Rec. & F1 \\
          \midrule 
          % \multirow{2}*{TACRED} 
          & LaVeEntail$^{+}$  & 63.20$^{*}$ &  59.80$^{*}$ & \multicolumn{1}{c}{\text{61.40$^{*}$}} \\
          & Clean-LaVe$^{+}$  &  72.60 & 59.98 & 65.59\small{$\uparrow4.1$} \\
          \midrule
        \end{tabular}
    }
    \caption{The results of using original constraints defined in LaVeEntail. Upper $+$ means full constraints and the results of LaVeEntail marked with * are retrieved from the original paper. }
    \label{tab:full_constraints}
\end{table}

\subsection{Cross-lingual Silver Standard Data} 
We further explore the potential of cross-lingual silver standard data. 
We combine the silver data from the source language (i.g., English) with the silver data from the target language and fine-tune the TE model.
Note that we do not use any labeled data from the source language as well as the target language. 
Results show that by using cross-lingual silver data from English, Clean-LaVe can further improve 0.2\% - 1.7\%.
%The improvement on the Italian language is relatively limited.
%This can be attributed to the larger quantity of silver data available for Italian compared to other languages. 
%Consequently, the performance gain achieved by using only the Italian silver data is already quite substantial.

\begin{table}[htbp]
    \label{tab:gen}
    \centering
    \resizebox{0.45\textwidth}{!}{
        \begin{tabular}[H]{p{4cm}<{\centering} | c | c | c}
             %\midrule
             %\multicolumn{4}{c}{Zero-shot Cross-lingual RE} \\
             \midrule
              & Italian & Polish & Korean \\
          \midrule
          LaVeEntail & 39.96 & 37.84 & 44.30 \\
          Clean-LaVe  & 55.09 & 52.99 & 59.41 \\
          Clean-LaVe {+ En Silver}  &  56.81\small{$\uparrow1.72$} & 53.23\small{$\uparrow0.24$} & 60.74\small{$\uparrow1.33$} \\
          \midrule 
        \end{tabular}
        }
    \caption{The average of F1 scores in 3 runs of Clean-LaVe on the zero-shot cross-lingual RE task. }
    \label{tab:gen}
\end{table}

\subsection{Hyper-parameter Analysis}

As Section \ref{sec:data selector} mentioned, Class-Aware Data Selector introduces two hyper-parameters to control the selection. $\eta$ controls the number of clean data and $m$ controls the number of samples from diverse classes. We analyse these hyper-parameters on 1\% development set on each dataset.

\begin{figure}[t]
	\centering
	\subfigure[RE task]{\label{fig:ppl-yelp2015}
		\includegraphics[width=0.48\textwidth]{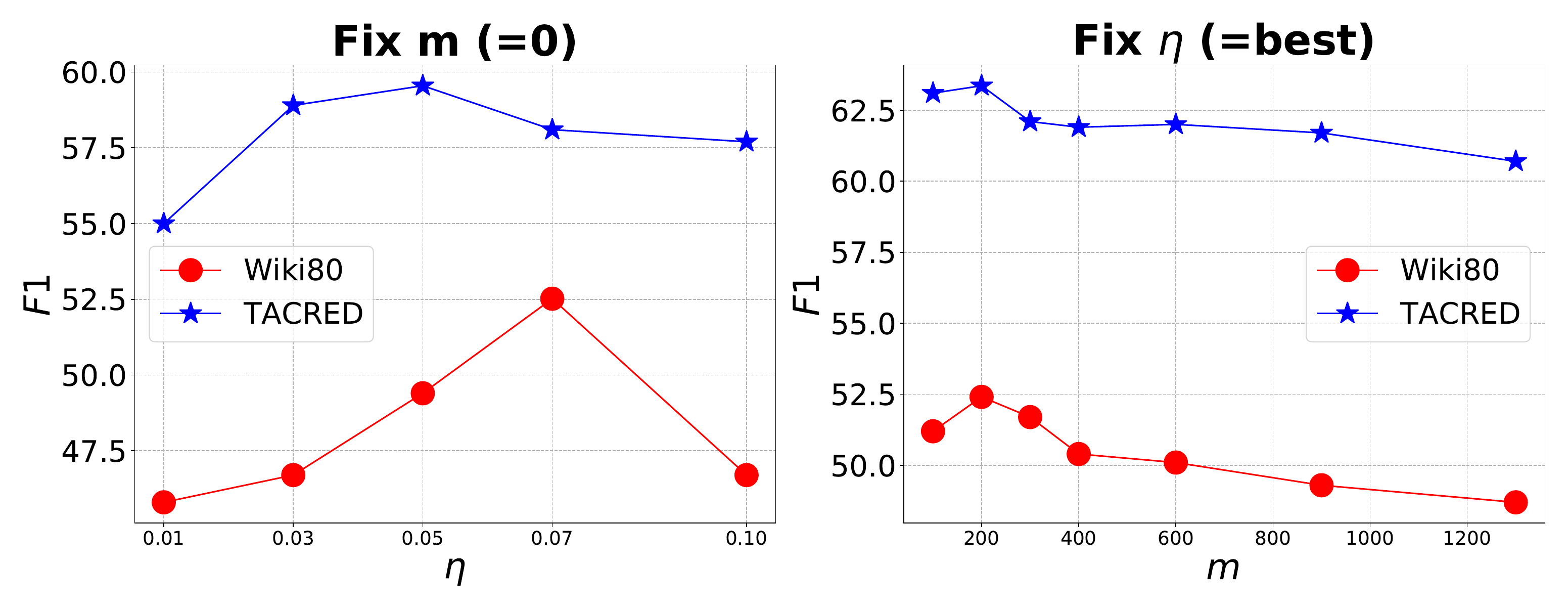}
	}
	\subfigure[EAC task]{\label{fig:ppl-yelp2016}
		\includegraphics[width=0.48\textwidth]{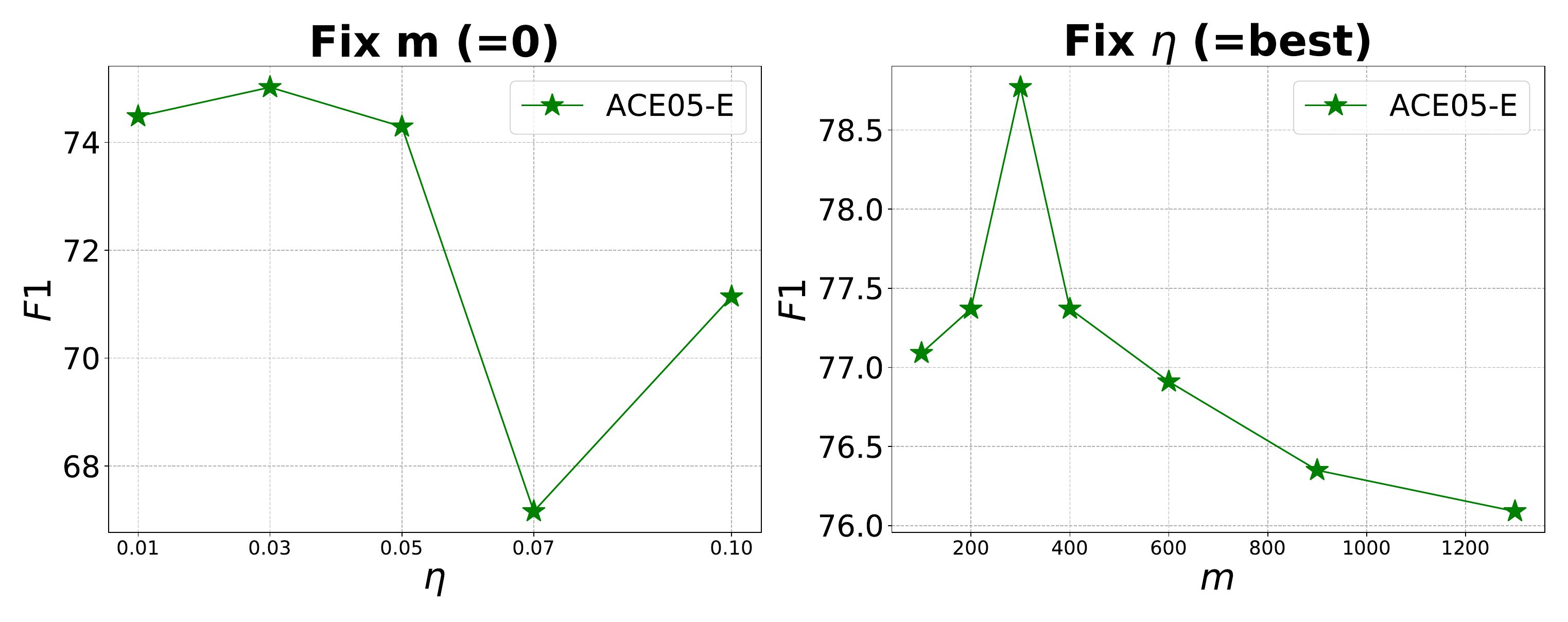}
	}
	\subfigure[Cross-lingual RE task]{\label{fig:ppl-r}
		\includegraphics[width=0.48\textwidth]{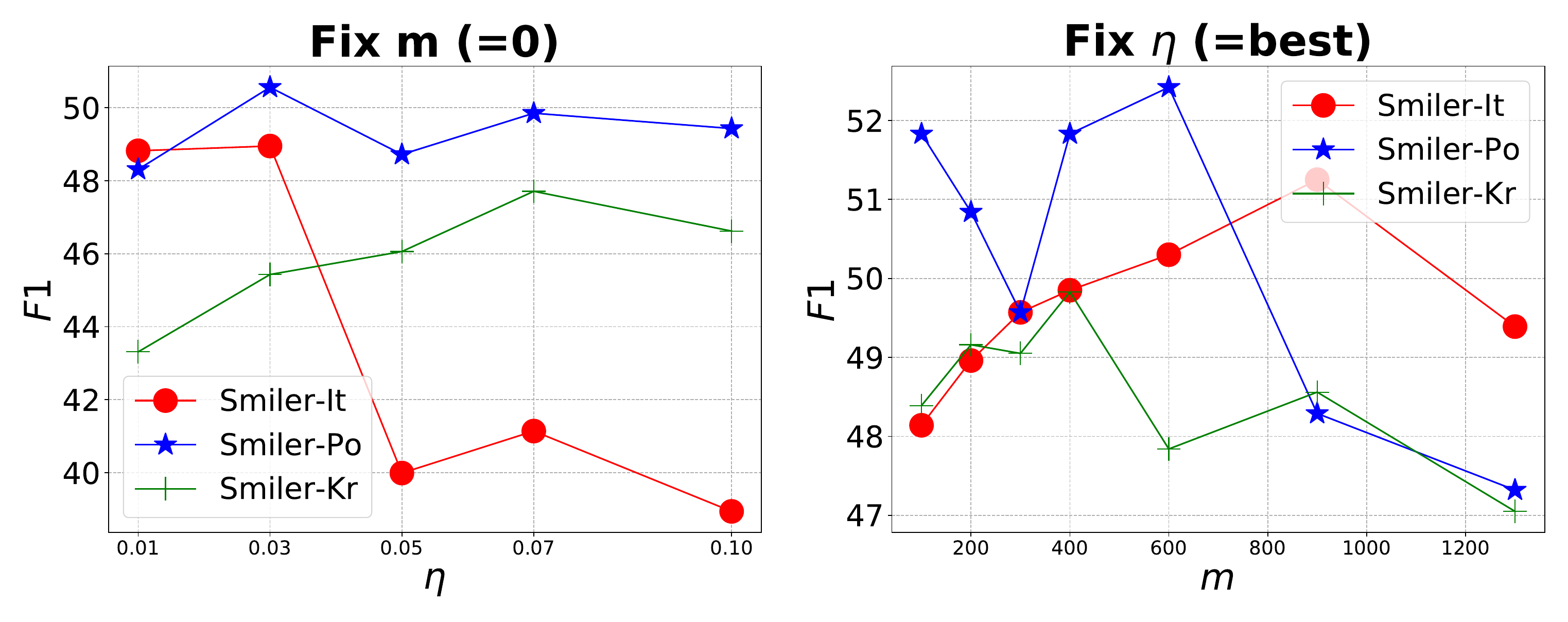}
	}
	\vspace{-0.1in}
	\caption{
 Results of different $\eta$ and $m$. 
 %Results of finetuning hyper-parameters ($\eta$ and $m$) in Class-Aware Data Selector module on 1\% development set for different datasets. We maintain one hyperparameter at a constant value while tuning the other.
	}	
 \label{fig:ppl}
\end{figure}

\noindent \textbf{Clean Data Selection Proportion $\eta$.}
We select $\eta \cdot |D_{silver}|$ data to finetune the TE model. 
The search range for $\eta$ is $[0.01,0.03,\cdots,0.1]$, while keeping another hyper-parameter, $m$, fixed at 0 to eliminate its influence.
As shown in Figure \ref{fig:ppl} (left column), with the increase of parameter $\eta$, the performance of the Clean-LaVe method increases first and then decreases. 
When $\eta$ is too small, although $D_{clean}$ has a low noise level, it only contains a few samples and classes, thus the model performance is barely satisfactory. 
When $\eta$ is too large, it easily involves too many noisy samples, thus deteriorating performance.

\noindent \textbf{Diversity Number $m$.}
We evaluate the effects of hyper-parameter $m$ which controls the number of samples from diverse classes. 
The search range for $m$ is $[100,200,\cdots,1300]$, while we maintain the value of $\eta$ fixed at the best value found during previous tuning.
As shown in Figure \ref{fig:ppl} (right column), with the increase of $m$, the performance generally increases since when $m$ is too large, it easily involves too many noisy samples, thus deteriorating performance.

\section{Conclusion}
We propose a framework named Clean-LaVe to first detect a small amount of clean data from silver standard data and then use them to finetune the pre-trained model. 
We propose a Iteratively Weighted Negative Learning algorithm and Class-Aware Data Selector in clean data detection process to alleviate the imbalanced issue and to broaden the range of classes during selection. 
The experimental results demonstrate the effectiveness of our proposed method.
%on various zero-shot classification tasks in information extraction area. 
%Furthermore, by incorporating extra silver standard data of different distributions, Clean-LaVe can be further improved. 
%Clean-LaVe can also be applied to other zero-shot tasks, such as zero-shot cross-lingual RE, zero-shot event detection, and zero-shot event argument classification.

\clearpage

\section{Bibliographical References}
\bibliographystyle{lrec-coling2024-natbib}
\bibliography{custom}

\section{Language Resource References}
\label{lr:ref}
\bibliographystylelanguageresource{lrec-coling2024-natbib}
\bibliographylanguageresource{languageresource}

% \clearpage
% \appendix
% \input{appendix}

\end{document}